\documentclass[conference]{IEEEtran}
\IEEEoverridecommandlockouts
\def\IEEEtitletopspace{18pt}

\usepackage{amsmath,amssymb,amsfonts}
\usepackage{algorithmic}
\usepackage{graphicx}
\usepackage{textcomp}
\usepackage{xcolor}
\usepackage{booktabs}
\usepackage{subfig}

\usepackage{multirow} 
\usepackage{url} 
\usepackage{xspace}

\newcommand{\ours}       {{HACTS}\xspace}
    
\begin{document}

\title{HACTS: a Human-As-Copilot Teleoperation System for Robot Learning}

\author{%
 Zhiyuan Xu$^{1,*}$, Yinuo Zhao$^{1,2,*}$, Kun Wu$^{1,*}$, Ning Liu$^{1}$, \\
 Junjie Ji$^{1}$, Zhengping Che$^{1}$, Chi Harold Liu$^{2}$, Jian Tang$^{1,\dagger}$\\
 $^1$Beijing Innovation Center of Humanoid Robotics\\
 $^2$Beijing Institute of Technology\\
 \thanks{$^{*}$Authors with equal contribution.
 \{Eric.Xu, Gongda.Wu\}@x-humanoid.com, ynzhao@bit.edu.cn. 
 This work was done during Yinuo Zhao's internship at Beijing Innovation Center of Humanoid Robotics. $^\dagger$Corresponding author: Jian Tang.}
}


\maketitle

 \begin{abstract}
Teleoperation is essential for autonomous robot learning, especially in manipulation tasks that require human demonstrations or corrections. However, most existing systems only offer unilateral robot control and lack the ability to synchronize the robot’s status with the teleoperation hardware, preventing real-time, flexible intervention. In this work, we introduce HACTS (Human-As-Copilot Teleoperation System), a novel system that establishes bilateral, real-time joint synchronization between a robot arm and teleoperation hardware. This simple yet effective feedback mechanism, akin to a steering wheel in autonomous vehicles, enables the human copilot to intervene seamlessly while collecting action-correction data for future learning. Implemented using 3D-printed components and low-cost, off-the-shelf motors, HACTS is both accessible and scalable. Our experiments show that HACTS significantly enhances performance in imitation learning (IL) and reinforcement learning (RL) tasks, boosting IL recovery capabilities and data efficiency, and facilitating human-in-the-loop RL. HACTS paves the way for more effective and interactive human-robot collaboration and data-collection, advancing the capabilities of robot manipulation.
\end{abstract} 

%
%
%

\section{Introduction}
Teleoperation plays a vital role in the development of robot learning, particularly in manipulation algorithms that rely on human demonstration, such as Vision-Language-Action (VLA) and Human-in-the-Loop Reinforcement Learning (HITL RL). These algorithms have shown significant success in recent years, enabling robots to perform complex tasks by leveraging human demonstration data. For instance, VLA models combine vision, language, and action to enhance robot autonomy, while HITL RL allows for online learning through human intervention. The integration of teleoperation in these frameworks has opened new avenues for enhancing robot capabilities and efficiency.

\begin{figure}[tbp]
\centerline{\includegraphics[width=1\linewidth]{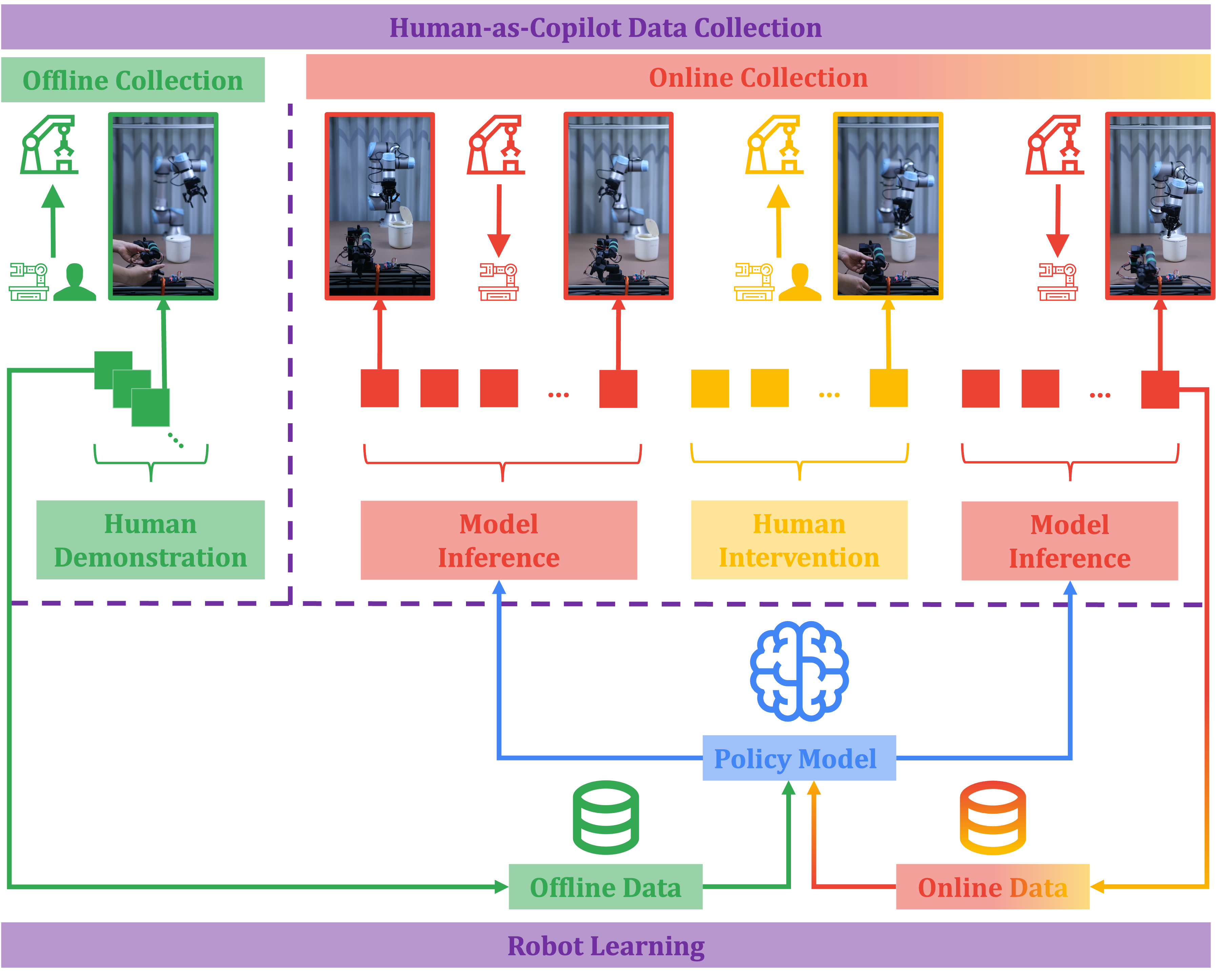}}
\caption{
We introduce HACTS, a Human-As-Copilot Teleoperation System that supports bilateral joint control for robot learning. During offline data collection, HACTS serves as a kinematically equivalent controller to collect precise human demonstration data. During online collection, its reverse synchronization feature enables seamless human intervention, providing online action-correction data for policy model learning. HACTS thus enhances both imitation learning and reinforcement learning methods.}
\label{fig}
\end{figure}

However, current teleoperation systems are limited by their hardware, such as VR~\cite{khazatsky2024droid}, exoskeletons~\cite{fang2024airexo}, and motion-capture technologies~\cite{darvish2023teleoperation}, which are designed for unilateral control. In these systems, the human operator can send commands to the robot but receives no real-time feedback, a limitation when the robot is performing autonomous tasks and requires human intervention.  Without feedback, the robot may make unintended, sudden movements, disrupting task continuity and performance. 
This is akin to a driver of an autonomous vehicle attempting to intervene while the vehicle is in motion—without a mechanism to sync the vehicle’s wheel movement with the driver’s input, the intervention could lead to a loss of control~\cite{zhang2020toward}.
Recent work~\cite{kobayashi2025alpha, liu2025factr, toedtheide2023force} has proposed bilateral control systems that integrate tactile and force feedback into teleoperation hardware. However, these systems require the follower device to generate precise force or tactile sensing information—capabilities that are not widely available on most robotic platforms.
Thus, there is a clear need for a teleoperation system offering bilateral joint synchronization between the teleoperation system and the robot, ensuring smooth transitions and more reliable interventions, as well as being compatible with a broad range of robots. 

To address this limitation, in this paper, we propose HACTS (Human-As-Copilot Teleoperation System), a low-cost but effective solution that enables \textbf{bilateral, real-time joint synchronization} between the robot arm and the teleoperation hardware. The HACTS system is implemented using only 3D-printed components and low-cost, off-the-shelf motors, making it both affordable and scalable. This simple yet effective hardware setup allows for continuous joint feedback from the robot to the teleoperation hardware, enabling more intuitive and controlled intervention when required. It offers a robust, seamless connection between the operator and the robot arm, making it ideal for human-in-the-loop applications.

We design a series of experiments to demonstrate that HACTS facilitates the collection of action-correction data, enhancing the performance of imitation learning algorithms. We show that policy learning models, when trained with data collected via HACTS, exhibit improved failure correction capabilities, as well as enhanced data efficiency in improving generalization under both static and dynamic scenarios. Furthermore, we demonstrate that HACTS supports more complex HITL RL setups, enabling the robot to adapt and learn from both human intervention and autonomous actions in a seamless loop. These advancements open up new possibilities for more effective and interactive human-robot collaboration, ultimately enhancing robot learning systems. In summary, the main contributions of this work are threefold:
\begin{itemize}
    \item We introduce a novel teleoperation system, HACTS, that enables real-time, bilateral joint synchronization for more intuitive human-robot interaction.
    \item We demonstrate that HACTS improves imitation learning performance, enhancing failure recovery and data efficiency in IL models.
    \item We show that HACTS facilitates more complex HITL RL, expanding the capabilities of online robot learning systems.
\end{itemize}
\section{Related Works}

\subsection{Teleoperation for robot manipulation learning}

Despite recent advances in robot manipulation driven by AI, fully autonomous solutions remain far from achieving socially and physically competent robot behaviors. This is why teleoperated robots, serving as physical avatars for human workers on-site, present a more practical solution~\cite{darvish2023teleoperation, tong2024advancements}. 
Furthermore, with the rise of learning-based, data-driven approaches in robotic control, teleoperation has become an essential system for gathering manipulation data~\cite{wu2024robomind, khazatsky2024droid}.
Traditional teleoperation systems, such as VR~\cite{zhang2018deep}, joystick controller~\cite{Kinova}, Phone-based~\cite{mandlekar2018roboturk}, and 3D Spacemouse~\cite {luo2024precise}, can only control the end-effector pose of the robot arm that heavily relies on IK/FK and motion planning, which limits the expression of human operators.
More recent approaches have highlighted the potential of low-cost, kinematically equivalent controllers, such as ALOHA~\cite{zhao2023learning} and Gello~\cite{wu2024gello}, which enable more dexterous and precise control, providing valuable data for robot learning. However, these systems have primarily focused on unilateral control, restricting the full potential of the hardware. 
Several works have attempted to develop bilateral control systems for robotic teleoperation~\cite{toedtheide2023force, liu2025factr, kobayashi2025alpha}, focusing on providing force or tactile feedback to human operators. 
However, these complex bilateral systems require expensive follower hardware and carefully designed algorithms to handle force or tactile feedback, which are not accessible on most robotic platforms like the UR5.
Although the Bi-ACT~\cite{kobayashi2025alpha} system could operate without force feedback, it has only been tested on identical leader-follower setups, and its applicability to larger, more common robotic platforms remains unknown.

In this work, we propose a similar low-cost, kinematically equivalent device that incorporates bilateral control, enabling humans to act as copilots and intervene when needed. We demonstrate that even a simple bilateral position synchronization, which is compatible with almost all robotic systems, can significantly enhance control precision and data collection capabilities. Our approach highlights the potential of more accessible, cost-effective teleoperation systems for improving robot learning and human-robot interaction.
Concurrent work RoboCopilot~\cite{wu2025robocopilot} introduces a similar human-in-the-loop teleoperation setup, whereas our work further demonstrates a reinforcement learning implementation based on the proposed HACTS framework.

\subsection{Learning from human demonstration and correction}

Learning from Demonstration (LfD) is a key paradigm in robot learning, enabling robots to acquire new skills by learning from human demonstrations rather than through explicit programming. Among the notable advancements in LfD are Vision-Language-Action (VLA) models~\cite{ma2024survey, black2024pi_0, wen2025tinyvla, wen2025dexvla}. 
%
%
These approaches facilitate learning multimodal human demonstrations, increasingly leveraging larger model architectures with more parameters. As a result, robots can benefit from richer contextual information, which improves task performance and aligns with the scaling laws observed in large language models. 
However, the process of collecting human demonstrations can be expensive and time-consuming, with previous work primarily focusing on success trajectories~\cite{wu2024robomind, khazatsky2024droid}. 
In contrast, our HACTS, with its simple bilateral control, allows for human intervention and generates more diverse types of data. Recent research has underscored the importance of failure data in training~\cite{wu2025swbt}, and we believe our system can significantly contribute to advancing this line of work by supporting the collection of such critical data for more robust learning.

The other group of methods that can benefit from {\ours} is Human-In-The-Loop (HITL) learning, which refines robot behavior through interactive feedback and enables policy adjustments based on human corrections. 
HITL methods can be categorized into passive human intervention~\cite{ross2011reduction, luo2023human}, where humans assist periodically or as determined by functions, and active human intervention~\cite{kelly2019hg, NEURIPS2023_f57ffe47, luo2024precise}, where humans decide when to intervene. 
Recently, approaches such as HACO~\cite{NEURIPS2023_f57ffe47} and HIL-SERL~\cite{luo2024precise} leverage active human assistance to enhance RL exploration and ensure safety during training, improving sample efficiency and exhibiting robust behaviors compared to classical RL and imitation methods like HG-Dagger~\cite{kelly2019hg} and IWR~\cite{mandlekar2020human}. 
However, these approaches require agile hardware platforms for flexible, timely intervention. 
In contrast to previous teleoperation systems, our HACTS offers more dexterous, precise control during human intervention, which we believe will inspire further innovation in HITL RL, especially for real-world online robot learning.
\section{System Design}

\subsection{Overview}

\begin{figure}[tp]
\centerline{\includegraphics[width=1.0\linewidth]{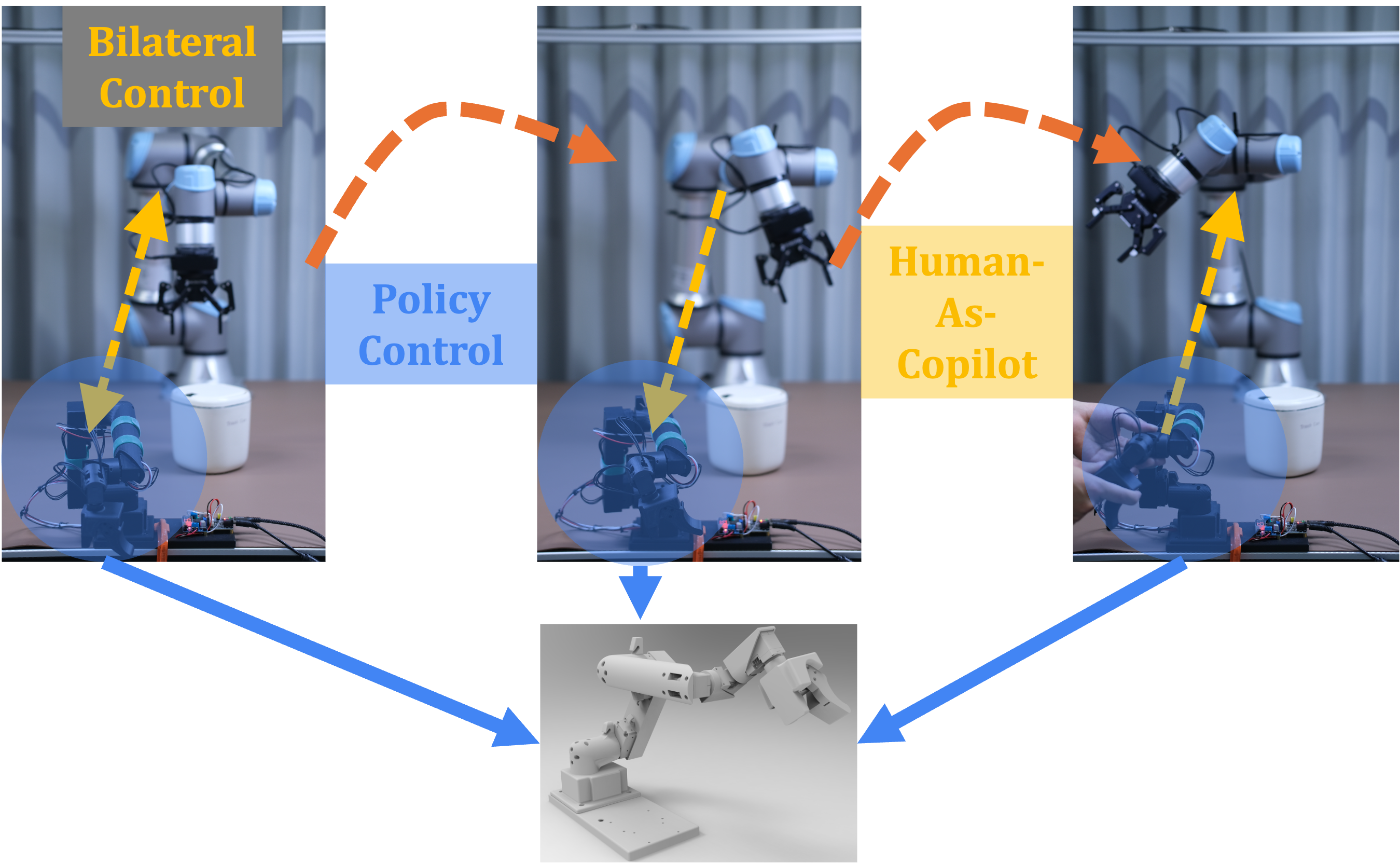}}
\caption{The overview usage of HACTS.}
\label{fig:system}
\end{figure}


We illustrate the key advancements of HACTS in Fig.~\ref{fig:system}. HACTS operates as a leader-follower system with bilateral position control, enabling both leader-to-follower and follower-to-leader synchronization. In leader-to-follower synchronization, HACTS functions similarly to systems like ALOHA~\cite{zhao2023learning} and Gello~\cite{wu2024gello}, facilitating the rapid and precise collection of offline demonstration data. Conversely, in follower-to-leader synchronization, where model inference drives the robot arm, HACTS enables the human operator to act as a copilot, seamlessly intervening when necessary to provide online action-correction data. Below, we outline key hardware and software design principles in HACTS. 

\subsection{Hardware Design}

\noindent \textbf{Low-Cost}.
One of our primary goals is to demonstrate that a capable teleoperation system can be built at an affordable price, making it accessible for large-scale deployment. To achieve this, we opted for servo motors rather than more expensive robotic modules, which are commonly used in more complex systems~\cite{darvish2023teleoperation}. The use of servo motors allows us to implement both passive position reading and active position control at a significantly lower cost. Previous works, including ALOHA~\cite{zhao2023learning}, Gello~\cite{wu2024gello}, and FACTR~\cite{liu2025factr}, have shown the impressive performance and stability of low-cost DYNAMIXEL motors, which further supports our design choice.

\noindent \textbf{Lightweight}.
Given our selection of low-cost servo motors, it was crucial to minimize the overall weight of the system to ensure that the motors could adequately support the hardware. This is particularly important for the first and second motors, which bear most of the force during active control. Specifically, we use the DYNAMIXEL XL430-W250-T motors for the first three links, where higher torque is required, and the XL330-M288-T motors for the gripper and rest of the links, which are lighter and more affordable. Additionally, we use 3D-printed PLA materials for the frame and structural components, further reducing the weight of the whole system.

\noindent \textbf{Easy-to-Use}.
Building upon prior work that demonstrated the effectiveness of miniature~\cite{wu2024gello}, kinematically equivalent devices for teleoperation, we aimed to create hardware that is intuitive and user-friendly, even for non-experts. This simplicity is key to ensuring that the system can be scaled for broader use, i.e., quickly adapt to a new robot. The kinematically equivalent structure is derived from the Denavit–Hartenberg parameters, with the lengths scaled by a factor to maintain the system's usability while optimizing for lightweight design. Note that these components can easily be extended to other robot arms since we only need to synchronize position information, which is generally available for most robotic arms. 

\subsection{Software Design}

The software stack for this system adopts a streamlined architecture to facilitate efficient hardware-robot communication. The DYNAMIXEL API serves as the primary interface for direct motor position monitoring and adjustment. Following an initial offset calibration procedure, the processed control joint values are transmitted to the robot controller through this interface. For reverse synchronization operations, the system implements a bidirectional data flow: robot position data undergoes reverse offset compensation to restore original positional values before being relayed to the motor units, thereby maintaining system-wide synchronization. An external foot pedal functions as the manual override interface, enabling operational mode switching between autonomous operation and human intervention states.

\section{Experiments}

We evaluate the efficacy of HACTS by addressing four key research questions:
\textbf{Q1}: Can human intervention data collected via the HACTS system serve as a viable substitute for an equivalent volume of direct teleoperation data?
\textbf{Q2}: Can the action-correction data in HACTS further improve the model's success rate under in-distribution (ID) scenarios?
\textbf{Q3}: Can the action-correction data in HACTS enhance the model's generalization capability for out-of-distribution (OOD) scenarios?
\textbf{Q4}: Can HACTS further support human-in-the-loop online reinforcement learning? 

To investigate these questions, we developed real-world robotic arm environments, designed a range of manipulation tasks, and established diverse experimental setups incorporating the two primary learning paradigms: imitation learning (IL) and reinforcement learning (RL).



\subsection{Hardware Setup}


\begin{figure}[tp]
\centerline{\includegraphics[width=1.0\linewidth]{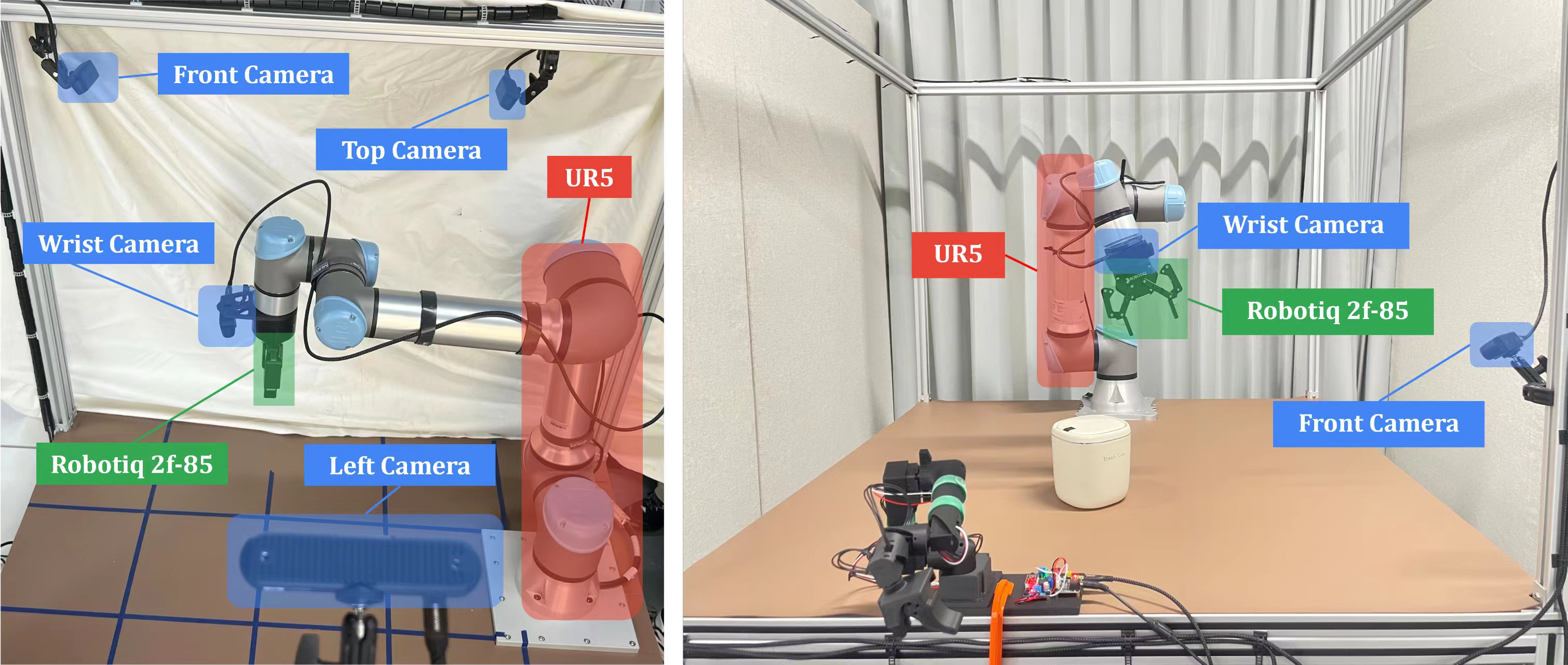}}
\caption{Experimental setup: (left) for imitation learning and (right) for reinforcement learning.}
\label{fig:workstations}
\end{figure}

As shown in Fig.~\ref{fig:workstations}, we built HACTS on two workstations, each equipped with one wrist-mounted 336 camera, and one or multiple 336L cameras.
As illustrated in Fig.~\ref{fig:workstations}, these cameras are positioned at different angles to capture a comprehensive set of visual inputs from varying perspectives. 
All components—including the cameras, the UR5 robot with Robotiq 2f-85 gripper, the HACTS system, and the policy model—are integrated into a single RTX 4090 workstation, ensuring robust computational performance for real-time data processing.
To demonstrate the feasibility and effectiveness of HACTS, we implemented the hardware using the UR5 kinematic model, and reused parts of the structure design in~\cite{wu2024gello} but modified them to fit the XL430 motors.
The whole cost for HACTS hardware is lower than $\$300$, and we listed all components in Tab.~\ref{tab:bill}.

\begin{table}[htp]
    \centering
    \caption{The bill of materials of HACTS hardware for UR5.}
    \begin{tabular}{c|cc}
    \toprule
        Components & Quantity & Cost\\
        Dynamixel XL430-W250-T & 2 & \$100 \\
        Dynamixel XL330-M288-T & 5 & \$120 \\
        12V 5A Power Supply & 1 & \$10 \\
        WaveShare Bus Servo Adapter & 1 & \$5 \\
        Voltage Reducer & 1 & \$2 \\
        3D Printed Plastic & - & \$10 \\
    \midrule
        Total & - & \$247 \\
    \bottomrule
    \end{tabular}
    \label{tab:bill}
\end{table}

\subsection{Experimental Setup on Imitation Learning}

\textbf{Task Design.}
To evaluate the use of the HACTS within the framework of imitation learning, we designed three real-world robotic tasks. 
These tasks required the robotic arm to learn and execute complex manipulation actions, such as grasping, placing, pressing, flipping, and precisely identifying the location of objects.
The objects used in these tasks are shown in Fig.~\ref{fig:act_tasks}. 
\textbf{OpenBox}: Accurately locate a switch of a box and press it to open the box. 
\textbf{SteamBun}: Grasp a steamed bun from a plate and place it into a steamer. 
\textbf{UprightMug}: Flip an overturned mug upright, then place it back in a designated central position. 



\textbf{Algorithm Instances.}
We focus on popular visuomotor policy learning techniques, where the robot learns to perform tasks based on visual observations and proprioception. 
The input consists of the RGB image data and the robot proprioceptive states, while the output is the end-to-end robotic control signals, such as joint positions.
Imitation learning methods allow the robot to learn directly from human demonstrations, mimicking the demonstrated actions to perform similar tasks autonomously.
We employed two state-of-the-art imitation learning methods for our experiments:

\begin{itemize} 
    \item \textbf{Action Chunking Transformer (ACT)}~\cite{zhao2023learning}: ACT predicts a sequence of actions and leverages temporal ensemble technology to generate smooth and precise actions.
    \item \textbf{Diffusion Policy (DP)}~\cite{chi2023diffusion}: DP uses a diffusion model, which is expressive and can model multi-modal action patterns, to improve the success rates.
\end{itemize}

\textbf{Evaluation Protocol.}
For each task, we initially collected 100 expert trajectories through teleoperation.
From these, we randomly chose 50 expert trajectories to train the preliminary ACT and DP models, referred to as “pre-ACT$/$DP”. 
Additionally, we trained the ACT and DP models using the full set of 100 expert trajectories as the baseline models, referred to as “full-ACT$/$DP”.
%
Further, to answer the \textbf{Q1, Q2} and \textbf{Q3}, we designed four different experimental settings.
For each various experimental setting, we used HACTS to collect 50 human intervention trajectories (denoted as HACTS trajectories) with using the "pre-ACT$/$DP" to execute tasks.
Then we mixed these 50 HACTS trajectories with the initial 50 random expert trajectories to train the refined ACT and DP models, denoted as “HACTS-ACT$/$DP”.
The four experimental settings are as follows:

\begin{figure}[tp]
\centerline{\includegraphics[width=1.0\linewidth,trim=0 110 0 110,clip]{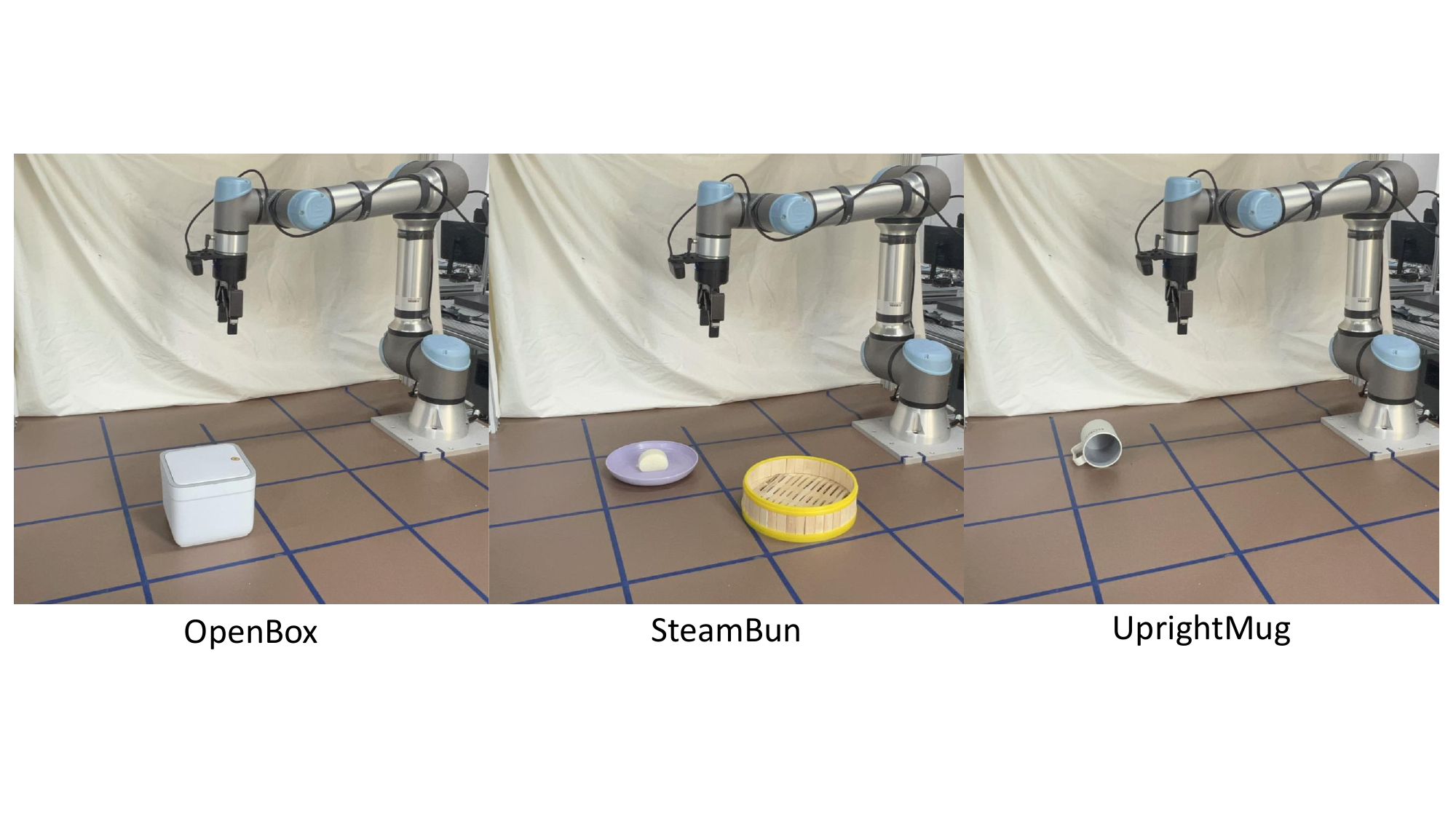}}
\caption{Three tasks for imitation learning experiments.}
\label{fig:act_tasks}
\end{figure}

\begin{itemize} 
    \item \textbf{Equal Amount Data Comparison (EADC)}: 
    This setting evaluates whether the human intervention data collected via \textbf{HACTS can serve as a substitute for an equivalent amount of expert teleoperation data.}
    During the inference of the preliminary models, regardless of whether the execution was successful, the data collector began recording human intervention data once the robotic arm reached a certain level of progress (e.g., moving from the initial pose to a certain height). 
    A total of 50 human intervention trajectories were collected in this setting.
    \item \textbf{Failure Correction in In-Distribution Scenarios (FCID)}: 
    This setting assesses the ability of \textbf{action-correction data to rectify failures in in-distribution scenarios.}
    Specifically, for scenarios where the objects are placed within the in-distribution region of the training dataset but the preliminary models still fail, we collected 50 action-correction trajectories to assess whether these data can help to correct the unsuccessful executions.
    \item \textbf{Generalization for Out-of-Distribution Static Scenarios (ODSS)}: 
    This setting examines whether \textbf{action-correction data can improve the model’s generalization for out-of-distribution static scenarios.}
    When the object placements are entirely different from those in the training dataset, there is a significant decline in the performance of the preliminary model. 
    In these cases, we collected 50 action-correction trajectories to evaluate whether these data can enhance the generalization ability of the models to different object positions.
    \item \textbf{Generalization for Out-of-Distribution Dynamic Scenarios (ODDS)}: 
    This setting involves scenarios where, during task execution, the target object is manually moved to a different position, leading to a sharp decline in the performance of the preliminary models. 
    We collected 50 action-correction trajectories to evaluate whether \textbf{these interventions can improve the model's robustness against dynamic disturbances.}
\end{itemize}

We employed ACT to conduct experiments for the first three experimental settings, while DP was utilized in the fourth setting. 
The reason for this choice is that the action chunking technique inherent in ACT combines actions over time dimension, making it inherently less suitable for dynamic tasks.
For each experiment, we selected the checkpoint with the lowest loss on the validation dataset and conducted 10 evaluation rollouts to assess the success rates of the models.



\begin{table}[tp]
    \centering
    \caption{Success rates of ACT on the first three experimental settings including EADC, FCID and ODSS for all tasks.}
    
    \begin{tabular}{c|c|ccc}
    \toprule
       Tasks & Setting & pre-ACT & full-ACT & HACTS-ACT \\
    \midrule
       \multirow{3}{*}{OpenBox} &  EADC & 50\%  &  60\% & 80\% \\
          &  FCID & 10\%  &  40\% & 50\% \\
          &  ODSS & 0\%  &  0\% & 30\% \\
    \midrule
       \multirow{3}{*}{SteamBun} &  EADC & 60\%  &  80\% & 90\% \\
          &  FCID & 20\%  &  60\% & 80\% \\
          &  ODSS & 0\%  &  0\% & 50\% \\
    \midrule
       \multirow{3}{*}{UprightMug} &  EADC & 40\%  &  60\% & 70\% \\
          &  FCID & 10\%  &  40\% & 70\% \\
          &  ODSS & 0\%  &  0\% & 40\% \\
   \bottomrule
    \end{tabular}
    \label{tab:act_results}
\end{table}

\subsection{Experimental Results on Imitation Learning}

Tab.~\ref{tab:act_results} presents the results of ACT on the three experimental settings including EADC, FCID, and ODSS for all three tasks, evaluated in terms of success rate.
Tab.~\ref{tab:dp_results} presents the results of the DP algorithm on the setting ODDS.

\textbf{Answer for Question 1}. 
We focused on the comparison between HACTS-ACT and full-ACT in the EADC setting in Tab.~\ref{tab:act_results}. 
%
Although trained on the same amount of data, HACTS-ACT achieved higher success rates than full-ACT in all three tasks, due to including more demonstrations in scenarios where pre-ACT models often failed, such as picking actions.
This indicates that the HACTS data can effectively replace expert teleoperation data without any performance drop. 
%

\textbf{Answer for Question 2}. 
We examined whether the performance of HACTS-ACT improves in the FCID setting. 
The results presented in Tab.~\ref{tab:act_results} reveal that HACTS-ACT not only significantly enhanced the success rates of pre-ACT across all three tasks, but also consistently outperformed full-ACT. 
For instance, in the UprightMug task, HACTS-ACT achieved a remarkable success rate of 70\%, compared to just 10\% for pre-ACT and 40\% for full-ACT. 
This improvement can be attributed to the ability of HACTS-ACT to specifically gather corrective data targeted at failed cases, whereas full-ACT merely increases the volume of data.

\textbf{Answer for Question 3}. 
We analyzed the performance of HACTS-ACT under the ODSS setting in Tab.~\ref{tab:act_results} and the ODDS setting in Tab.~\ref{tab:dp_results}. 
Notably, we found that the success rates of pre-ACT, pre-DP, full-ACT, and full-DP are all 0\% in these scenarios, indicating that they are unable to handle situations that are unseen in the training dataset. 
However, upon incorporating HACTS-generated action-correction data, both HACTS-ACT and HACTS-DP demonstrated the ability to effectively handle these OOD situations, thereby improving the model's generalization capabilities.


In summary, the experiments on imitation learning show that HACTS effectively supports IL methods by providing accurate and real-time feedback during human intervention. 
The results demonstrate that HACTS can improve model performance and learning efficiency by collecting valuable action-correction data and integrating them into the learning process.

\begin{table}[tp]
    \centering
    \caption{Success rates of DP on the experimental settings ODDS.}
    
    \begin{tabular}{c|c|ccc}
    \toprule
       Tasks & Setting & pre-DP & full-DP & HACTS-DP \\
    \midrule
       OpenBox &  ODDS & 0\%  & 0\% & 40\% \\
   \bottomrule
    \end{tabular}
    \label{tab:dp_results}
\end{table}

\subsection{Experimental Setup on Reinforcement Learning}
To investigate whether \ours~can support human-in-the-loop online learning, we implement \textbf{RLPD-HACTS}, which combines \ours~with RLPD~\cite{ball2023efficient}, a sample-efficient off-policy reinforcement learning algorithm that uses human demonstrations (expert data) and corrections (\ours~data) during online training. The training proceeds in three stages. In the first stage, similar to HIL-SERL~\cite{luo2024precise}, we train a binary classifier based on a pretrained ResNet-10~\cite{he2016deep} to process images from the front and wrist cameras. Using HACTS, we collect 200 positive data points indicating task completions and around 1{,}000 negative data points to train the reward classifier. In the second stage, we gather 20 trajectories of expert data via our system and use them to train a behavior cloning (BC) policy. The BC policy takes concatenated ResNet-10 embeddings, gripper states and proprioceptive states—defined by the relative 6D end-effector pose compared to its initial pose—as input, and the action space corresponds to the end-effector’s 6D movement and the gripper states. In the final stage, we load the pretrained actor network and fine-tune it with RLPD, collecting short, frequent human corrections via HACTS. The reward classifier is queried at every step during online training, returning 1 if the task is completed and 0 otherwise.

\begin{figure}[tbp]
\centerline{\includegraphics[width=0.8\linewidth]{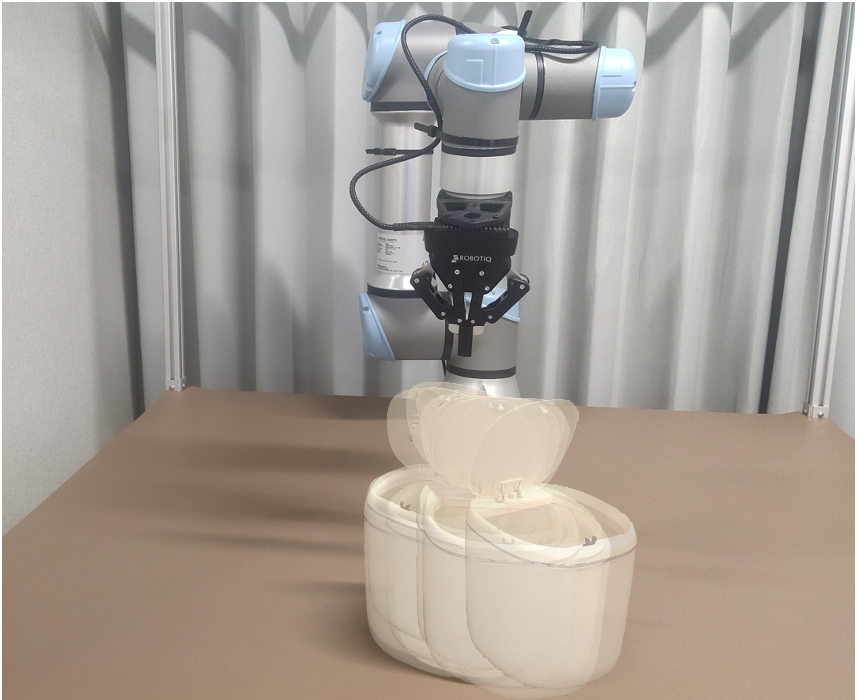}}
\caption{CloseBin task. The robot is required to close a trash bin that is randomly placed and oriented.}
\label{fig:rl_task}
\end{figure}

As shown in Fig.~\ref{fig:workstations}, we train and evaluate RLPD-HACTS on a UR5 workstation with our \ours~system, where the environment updates at 10 Hz. As shown in Fig.~\ref{fig:rl_task}, we design a CloseBin task where the robot is required to close a trash bin that is randomly placed and oriented. This task is challenging because precise motion control is required based on the bin cover’s position and orientation.

\subsection{Experimental Results on Reinforcement Learning}
We developed RLPD-HACTS by enhancing existing reinforcement learning methods with a human-in-the-loop, vision-based system. In this setup, a robot arm autonomously explores its environment but receives human assistance during challenging situations or when guidance is needed. This human input is seamlessly integrated into the learning process, enabling the RL algorithms to refine the control policy progressively by utilizing data from both robot and human-copilot actions.

\textbf{Answer for Question 4}. As shown in Tab.~\ref{tab:rl_result}, RLPD-HACTS achieves 80\% success rates after 10 minutes offline pretraining and 45 minutes online training. Compared with BC that trained only with human demonstrations, the policy trained by RL demonstrate shorter episode length due to the human assisted trial-and-error learning manner and the sparse reward settings. 


\begin{table}[t]
    \centering
    \caption{Results of RLPD-HACTS from offline BC pretraining followed by online finetuning. We report the success rates and episode lengths over 10 trials. }
    \scalebox{0.9}{
    \begin{tabular}{c|c|c|cc}
     \toprule
        \textbf{Tasks} & {\bf Training Time (mins)} & {\bf Success Rate (\%)} & {\bf Episode Length}\\ 
        \midrule
        CloseBin & 10 + 45   &50 $\rightarrow$ 80 & 32 $\rightarrow$ 19\\
   \bottomrule
    \end{tabular}}
    \label{tab:rl_result}
\end{table}

\begin{figure}[tbp]
\centerline{\includegraphics[width=1.0\linewidth]{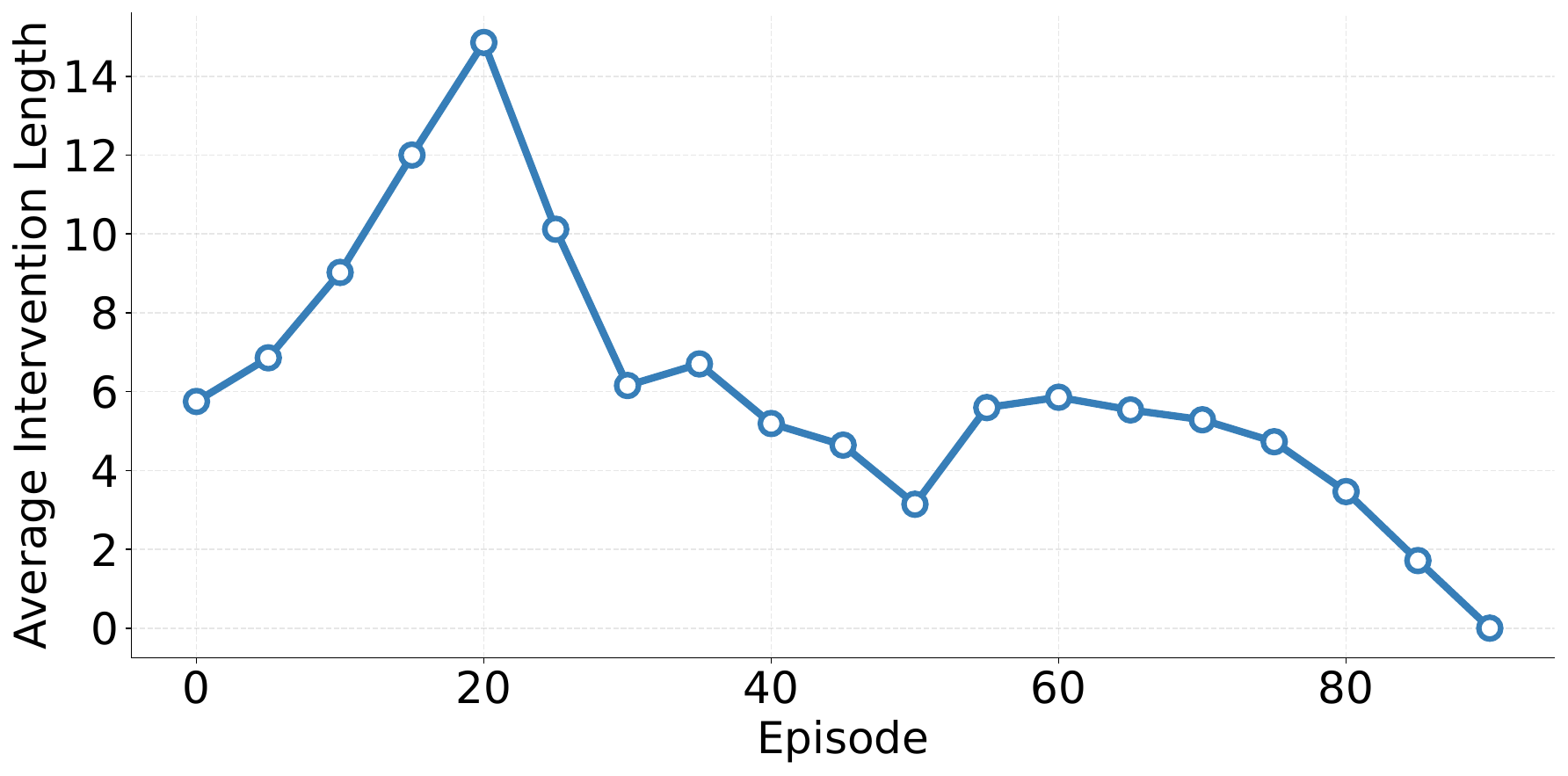}}
\caption{Average intervention length during RLPD-HACTS training in the CloseBin Task. Each episode terminates either after exceeding 200 steps or when the reward classifier outputs a logit value exceeding 0.85.}
\label{fig:human_intervention}
\end{figure}


In Fig.~\ref{fig:human_intervention}, we present the average human intervention length during online training, computed as the total number of intervention steps divided by the times of consecutive interventions. Initially, intervention length increases, reflecting a performance drop in RLPD-HACTS due to Q-network learning and exploration. During this phase, humans must provide long interventions to ensure success. As training progresses, the policy increasingly relies on its Q estimations, reducing the need for human input. Eventually, interventions drop below six steps, enabling fine-grained corrections and preventing overly aggressive closing strategies near the end of each episode.

As previously mentioned, \ours~can be seamlessly integrated into the human-in-the-loop reinforcement learning workflow. Compared to conventional teleoperation methods, such as a 3D spacemouse, \ours~offers superior dexterity and precision. Consequently, operators can intervene more effectively, guiding the robot with greater accuracy during complex tasks that require significant twisting or large-scale movements, as in many manipulation scenarios.



It is important to note that we only select some representative IL and RL method that directly uses \ours~for human intervention and data collection. Results from both IL and RL experiments show that \ours~can enhance existing algorithms by providing high-quality human copilot data, refining robot learning and improving task performance. We believe that with further model optimization and careful integration design of human feedback, HACTS will play an increasingly significant role in advancing robot learning systems.



    

\section{Conclusion and Discussion}

In this paper, we present HACTS (Human-As-Copilot Teleoperation System), a novel teleoperation system designed to enable more effective human-robot collaboration through bilateral joint synchronization. Unlike traditional teleoperation systems that provide unilateral control, HACTS allows for real-time feedback from the robot arm to the human operator, facilitating smoother interventions when needed. This bidirectional communication is essential for tasks involving imitation learning and human-in-the-loop reinforcement learning, where the operator can correct the robot’s autonomous actions, and valuable action-correction data can be collected simultaneously.

Through the implementation of low-cost, kinematically equivalent components, we demonstrate that HACTS offers a scalable and accessible solution for teleoperation, making it a promising tool for advanced robot learning applications. Our experiments showed that integrating action-correction data into training significantly enhances the performance of 
visuomotor policies, improving their recovery and data efficiency. Furthermore, the system supports more complex human-in-the-loop reinforcement learning, enabling robots to adapt more effectively in dynamic environments.

Overall, HACTS represents a significant step forward in the development of teleoperation systems for robot learning, with its ability to provide precise, bilateral control. By combining ease of use, affordability, and enhanced data collection capabilities, HACTS opens up new avenues for interactive, human-in-the-loop robot learning, setting the stage for future research and development in this field.

\bibliographystyle{unsrt}
\bibliography{0-ref}

\end{document}